\documentclass[letterpaper, 10pt, conference]{ieeeconf}
\IEEEoverridecommandlockouts
\usepackage{algorithm}
\usepackage{algorithmic}
\usepackage{amsfonts}
\usepackage{amsmath}
\usepackage{amssymb}
\usepackage{array}
\usepackage{bm}
\usepackage{breakcites}
\usepackage{color}
\usepackage{comment}
\usepackage{float}
\usepackage{gensymb}
\usepackage{graphicx}
\usepackage{multirow}
\usepackage[caption=false,font=footnotesize,subrefformat=parens,labelformat=parens]{subfig}
\usepackage{textcomp}
\usepackage{wrapfig}

\PassOptionsToPackage{hyphens}{url}\usepackage[breaklinks=true,hidelinks=true]{hyperref}

\graphicspath{{figures/}}

\DeclareMathOperator*{\argmax}{arg\,max}

\title{Learning the Next Best View for 3D Point Clouds via Topological Features}

\author{Christopher Collander, William J. Beksi, and Manfred Huber
\thanks{The authors are with the Department of Computer Science and 
        Engineering, University of Texas at Arlington, Arlington, TX, USA. 
        Emails: 
        christopher.collander@mavs.uta.edu,
        william.beksi@uta.edu,
        huber@cse.uta.edu.
        }
}

\begin{document}
\maketitle
\pagestyle{plain}

\begin{abstract}
In this paper, we introduce a reinforcement learning approach utilizing a novel
topology-based information gain metric for directing the next best view of a
noisy 3D sensor. The metric combines the disjoint sections of an observed
surface to focus on high-detail features such as holes and concave sections.
Experimental results show that our approach can aid in establishing the
placement of a robotic sensor to optimize the information provided by its
streaming point cloud data. Furthermore, a labeled dataset of 3D objects, a CAD
design for a custom robotic manipulator, and software for the transformation,
union, and registration of point clouds has been publicly released to the
research community.  
\end{abstract}


\section{Introduction}
\label{sec:introduction}
\begin{figure}
\centering
\includegraphics[width=0.4\textwidth]{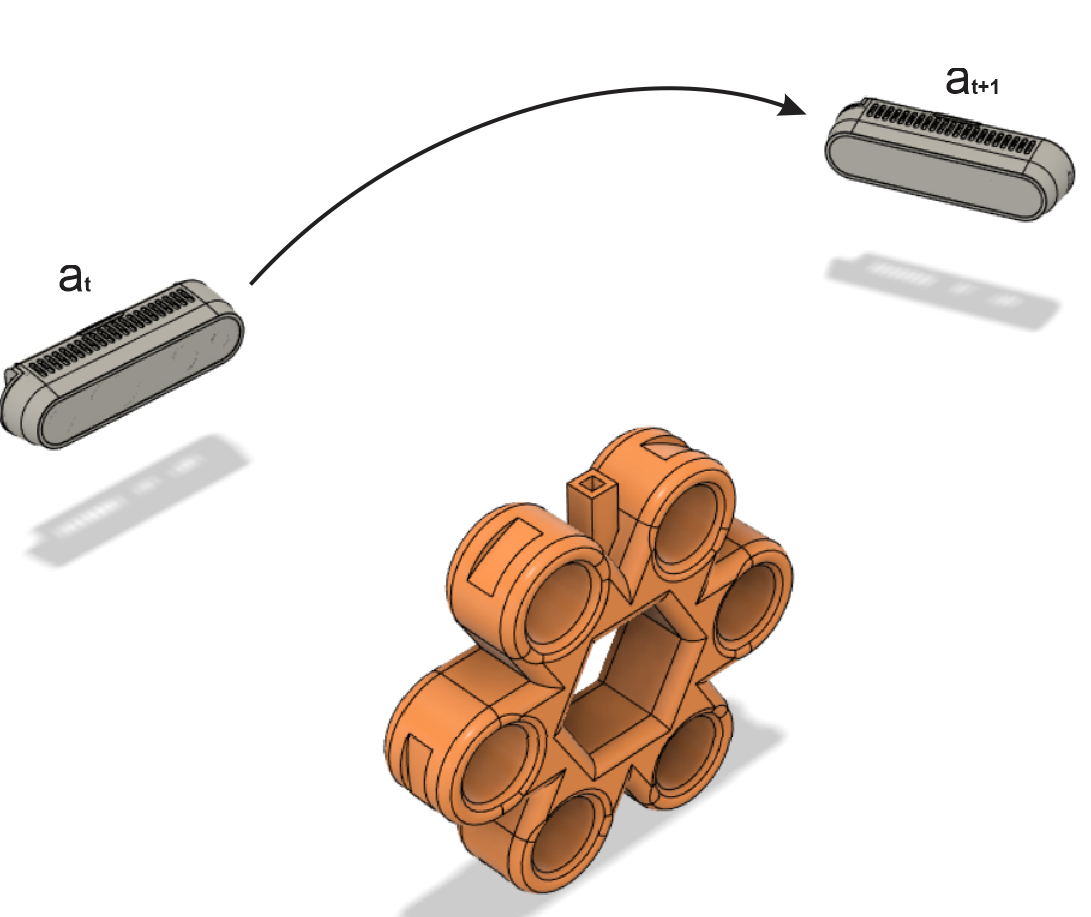}
\caption{The next best view (NBV) problem is the challenge of determining where 
to relocate a sensor, after an initial reading, to maximally increase the 
knowledge of the environment.}
\label{fig:nbv_example} 
\vspace{-4mm}
\end{figure}

Modern robots are equipped with a wide variety of sensors and actuators for
observing and interacting with the surrounding environment. This allows a robot
to carry out applications such as 3D reconstruction, object recognition,
grasping, and much more. To receive further clues about its surroundings, a
robot must move its sensor to another location and obtain new sensor values
\cite{aloimonos1988active,aloimonos1990purposive,chen2011active}. Several
factors constrain the acquisition of the next sensor view including the
kinematics of the robot, the field of view and range of the sensor, and
environmental obstructions such as occlusions \cite{maver1993occlusions} and
obstacles \cite{gonzalez2002navigation}. 
The challenge of obtaining a series of updated sensor placements is known as the
next best view (NBV) problem. Specifically, given a known sensor pose and an
information value, the next placement of the sensor should ensure that the value
obtained maximizes the `information gain' across the full action space of the
sensor, Fig.~\ref{fig:nbv_example}. Information gain is calculated between two
subsequent views, and the full gain of a sequence of views is the sum of the
individual consecutive gains. 

The definition of information gain is dependent upon the end goal of the robot.
In the context of reconstructing the surface of an object, information gain may
be defined as the metric of resolution \cite{hilton2000geometric}, mesh quality
\cite{kriegel2015efficient}, or volumetric representation
\cite{monica2018surfel}. 
For the task of object recognition, information gain can be formalized as the
fusion of successive object hypotheses \cite{paletta2000active}, a metric of
prediction surety \cite{wu20143d}, or a combination of probabilistic and
volumetric metrics \cite{daudelin2017adaptable}. In the case of picking and
transporting objects by robots in potentially cluttered environments,
information gain can be described by the Kullback-Leibler divergence
\cite{van2012maximally}, the utility of the sensor pose based on regions of
interest \cite{nieuwenhuisen2013mobile}, or the occupancy probability of a voxel
up to the most recent observation \cite{arruda2016active}. Enabling robots that
can act autonomously in dynamic, unstructured settings is a major challenge. In
such circumstances, learning to reconstruct, recognize, and manipulate
unfamiliar objects by choosing the NBV is an important capability.

In contrast to the aforementioned work, we propose to define information gain in
terms of persistent topological features modeled as rewards. The key insight of
this effort is that the incorporation of topological attributes can yield unique
knowledge with respect to the structure of point cloud data which is not
obtainable from other metrics. Our work makes the following contributions: (i) a
fully automated online deep reinforcement learning approach for achieving the
NBV on streaming 3D point cloud data; (ii) a novel and generalizable
topology-based information gain metric; (iii) the public release of a point
cloud dataset for seven exclusive objects in various labeled positions, an
open-source CAD design for a custom robot manipulator, and open-source software
for the transformation, union, and registration of point clouds given their
viewpoint coordinates.

The remainder of this paper is organized as follows. Related literature is
reviewed in Sec.~\ref{sec:related_work} and the background material for this
work is provided in Sec.~\ref{sec:background}. The statement of the problem
is given in Sec.~\ref{sec:problem_statement}. Our reinforcement learning
approach to directing the NBV via the computation of a topology-based
information gain metric is presented in Sec.~\ref{sec:approach}. The design
and results of our experiments are discussed and demonstrated in 
Sec.~\ref{sec:experiments}. In Sec.~\ref{sec:conclusion}, the paper is 
concluded and future work is mentioned. 

\vspace{-2mm}
\section{Related Work}
\label{sec:related_work}
Deciding the NBV is a fundamental area of research in active vision. It has over
thirty years of history and remains an open problem \cite{chen2008active}. We
distinguish between model-based NBV approaches, where the views can be planned
offline, and non-model-based algorithms where the NBV is selected at runtime
since no a priori knowledge about the target object is available. A
comprehensive survey on model-based and non-model-based methods is provided by
\cite{scott2003view}. Our approach develops a non-model-based scheme for finding
the NBV.  

The earliest characterization of the NBV used octrees to subdivide 3D space into
partitions \cite{connolly1985determination}. These partitions (empty, occupied,
or unseen) were used in two algorithms to determine the NBV. The first algorithm
(planetarium method) samples a sphere of views evenly across longitudes and
latitudes and compares the total area of unseen octree leaves. The second
algorithm (normal method) analyzes spatial faces that are common to both empty
and unseen partitions, and attempts to maximize the area of these faces in the
NBV. Both of these algorithms take multiple views to determine the NBV. In
comparison, our model predicts the NBV from only one initial view after
training. 

An automated surface acquisition system employed the NBV by partitioning a
viewpoint into seen and unseen views \cite{pito1996sensor,pito1999solution}. A
list of desired constraints for future views is constructed to increase the
expected quality and reduce the search state space. Referred to as the PS
algorithm, this method relies on determining which views constrain the ranging
rays to be colinear with observation rays of the object surface from the visual
sensor. Similar to our research, this work prioritizes views on the edge of the
initial view, resulting in some overlap between the views. In contrast to our
work, multiple views are evaluated before determining which is the NBV. This
method also only evaluates the NBV across a single dimension via the use of a
turntable to manipulate the object.



To determine which sensor poses have the highest probability for viewing missing
cells in a model, researchers have established a probabilistic framework for
expected information gain \cite{potthast2014probabilistic}. In their work, a
voxelized occupancy grid was used for calculations in non-simulated, real-world
scenarios of multiple objects cluttered on a small table. A path planning
algorithm utilizing rapidly-exploring random trees was proposed by researchers
for an aerial vehicle with a stereo camera to determine the NBV
\cite{bircher2016receding}. To explore a 3D environment, their algorithm chooses
views based on the amount of unmapped space. A voxelized occupancy grid was used
in a simulated environment for testing, and successful real-world tests were
also done with an unmanned aerial vehicle. Dissimilar to our work, both
\cite{potthast2014probabilistic} and \cite{bircher2016receding} attempt to
determine the NBV of a full environment while our research focuses on the NBV of
a single object.

In \cite{zeng2020pc}, point clouds from ShapeNet \cite{chang2015shapenet}
objects were examined for the NBV with a novel surface coverage metric. Unlike
our work, the research focuses on reconstructing an object in 3D space. In
contrast, our research concentrates on maximizing the realization of features in
the observed manifold. Specifically, we maximize the number of physical holes
and minimize both the number of holes due to missing points and the number of
connected components observed in the data. Another difference is that our work
is based on a single future view. On the contrary, \cite{zeng2020pc} takes 
multiple views while attempting to minimize how many views are required for 
reasonable surface coverage.

The research proposed in \cite{han2019deep} utilizes deep Q-learning for
multi-view planning with a reward value focused on depth-image inpainting.
Different from our work, this research does not convert the depth map into a
point cloud and therefore it does not calculate any graph-based values as we do.
Additionally, \cite{han2019deep} cannot find points outside of the dimensions of
the original depth image nor in opposing viewpoints. In our research, the pose
can be positioned in any location in spherical coordinates. \cite{han2019deep}
makes depth decisions of a full environment while our system works on the
observation of a single segmented object in space. 

\section{Background}
\label{sec:background}
This section provides the necessary background material on the computational
aspects of topological data analysis \cite{edelsbrunner2010computational} and
reinforcement learning \cite{sutton2018reinforcement} upon which our work is
based.

\subsection{Topological Data Analysis}
\label{subsec:topological_data_analysis}
Topological data analysis (TDA) refers to a collection of tools for extracting
topological features from data \cite{chazal2017introduction}. The main tool,
persistent homology, allows us to study homology (i.e., connected components,
holes, and higher dimensional analogs) at multiple scales
\cite{edelsbrunner2008persistent}. Concretely, it provides a basis to quantify
the evolution of the homology of a parameterized family of topological spaces. 


\begin{figure}
\includegraphics[width=1.0\columnwidth]{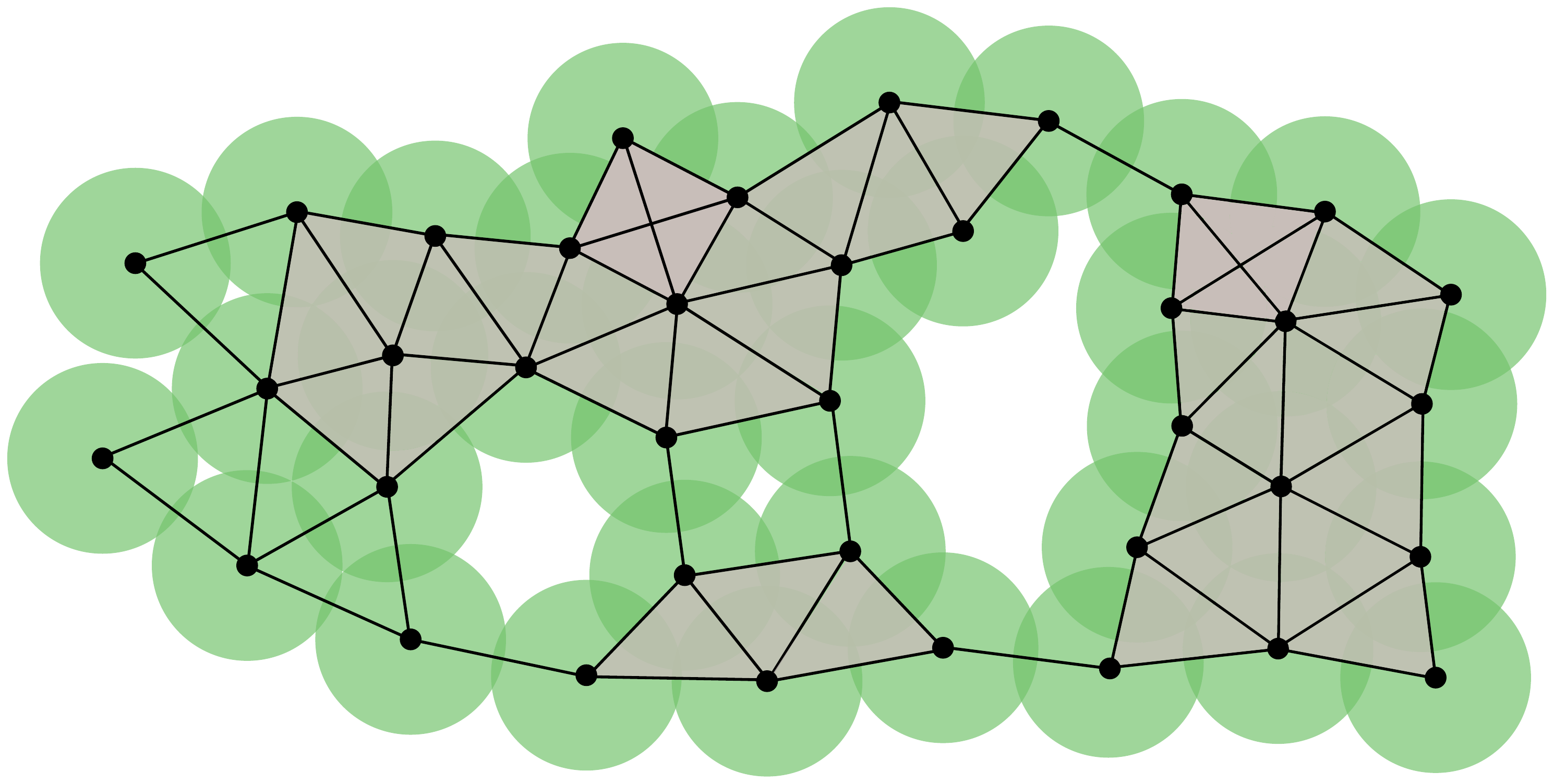}
\caption{0-simplices (vertices), 1-simplices (edges), and 2-simplices (shaded
triangles) can be glued together to form a Vietoris-Rips (VR) complex.}
\label{fig:vr_complex} 
\vspace{-4mm}
\end{figure}

\subsubsection{Simplicial Complex Representation}

The discrete space that we work in uses simplices as its building blocks. A
$k$-dimensional simplex $\sigma$ is the convex hull of $k+1$ affinely
independent vertices $v_0,\hdots,v_k \in \mathbb{R}^{n}$. For example,
0-simplices are vertices, 1-simplices are edges, and 2-simplices correspond to
triangles. Given a set of points $\{x_0,\hdots,x_{m-1}\} \in \mathbb{R}^{n}$ and
fixed radius $r$, the Vietoris-Rips (VR) complex is an abstract simplicial
complex (i.e., collection of simplices closed under the operation of taking
subsets) such that 
\begin{equation*} 
  K = \{\sigma \subset \{x_0,\hdots,x_{m-1}\} \mid dist(x_i,x_j) \le r, \forall x_i \ne x_j \in \sigma\}.  
\end{equation*} 
In this definition, $dist$ is the Euclidean distance function and the points of
$\sigma$ are pairwise within distance $r$ of each other,
Fig.~\ref{fig:vr_complex}.                                                                                                   


\subsubsection{Chains, Boundaries, and Cycles}
Given an abstract simplicial complex $K$, a $k$-chain is a subset of
$k$-simplices in $K$. A boundary, $\partial_k(\sigma)$, is a collection of
$(k-1)$ dimensional simplices and forms a $(k-1)$-chain. Taking the sum of the
boundaries of the simplices in a $k$-chain gives the boundary of the $k$-chain,
$\partial_k(c) = \sum_{\sigma \in c}\partial_k(\sigma)$. A homomorphism,
$\partial_k: C_k \rightarrow C_{k-1}$, is defined for each boundary operator and
the collection of boundary operators on the chain groups forms a chain complex,
\begin{equation*}
  \emptyset \rightarrow C_3 \xrightarrow{\partial_3} C_2
    \xrightarrow{\partial_2} C_1 \xrightarrow{\partial_1} C_0
    \xrightarrow{\partial_0} \emptyset.
\end{equation*}


\subsubsection{Homology Groups}

The homology groups comprise vector spaces where a subset yields a vector space
if every element is the sum of elements in the subset. A minimal generating set
serves as a basis for the space. The rank of the $k$th homology group is defined
as the $k$th Betti number of $K$, e.g. $\beta_k = \text{rank }H_k$ where
\begin{equation*} 
  \text{rank }H_k = \text{rank }Z_k - \text{rank }B_k.
\end{equation*} 
Only the Betti numbers for $0 \le k \le 2$ can be non-zero for complexes in 
$\mathbb{R}^3$. A non-bounding 0-cycle represents a set of components of $K$ 
where there is one basis element per component. Thus, $\beta_0$ is the number of 
components that make up $K$. A set of holes formed by $K$ is represented by a 
non-bounding 1-cycle. Hence, each hole can be expressed as a sum of holes in a 
basis and $\beta_1$ is the size of the basis.

\begin{figure}
\centering
\includegraphics[width=1.0\columnwidth]{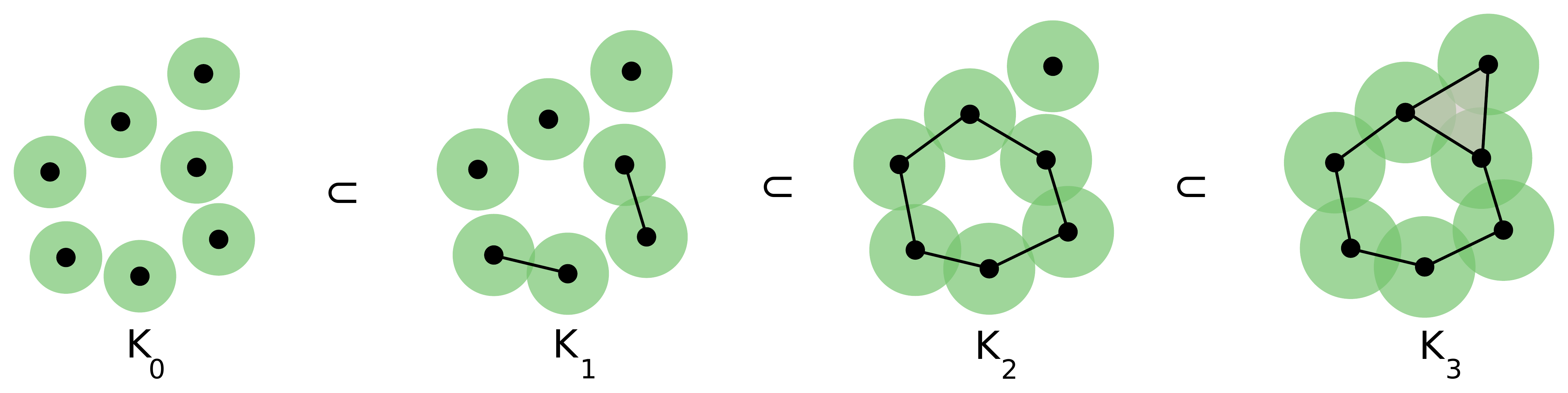}
\caption{A growing sequence of subcomplexes is known as a filtration.}
\label{fig:filtration}
\vspace{-4mm}
\end{figure}  

\subsubsection{Filtrations}
The evolutional growth of a complex of an increasing sequence of topological
spaces is called a filtration, Fig.~\ref{fig:filtration}. Concretely, if $f : X
\rightarrow \mathbb{R}$ is defined on a topological space $X$ then each sublevel
set $X_r = f^{-1}((-\infty, r])$ generates a topological space $X_r$ where $X_r
\subset X_r'$ and $r \le r'$. As the spatial neighborhood $r$ increases,
homological features are born (created) and can die (disappear). In this work,
we use filtrations of finite simplicial complexes in $\mathbb{R}^3$,
\begin{equation*} 
  \emptyset \subset K_0 \subset K_1 \subset \ldots \subset K_L = K_\infty, 
\end{equation*} where $L$ is a maximum threshold value for
constructing a complex.

\subsubsection{Persistent Homology}
Persistence measures the lifetime of topological attributes in a filtration.
More precisely, given the $l$th complex $K^{l}$ in a filtration, let $Z^{l}_{k}$
and $B^{l}_{k}$ be the corresponding $k$th cycle group and $k$th boundary group,
respectively. Then the persistent cycles in $K^{l}$ are obtained by factoring
the $k$th cycle group by the $k$th boundary group of $K^{l+p}$, $p$ complexes
later in the filtration. The p-persistent $k$th homology group of $K^{l}$ is
defined as
\begin{equation*}
  H^{l,p}_{k} = Z^{l}_{k} \,/\, (B^{l+p}_{k} \cap Z^{l}_{k}),
\end{equation*}
and the rank of $H^{l,p}_{k}$ is the p-persistent $k$th Betti number
$\beta^{l,p}_{k}$ of $K^{l}$.

\subsection{Reinforcement Learning}
\label{subsec:reinforcement_learning}
Reinforcement learning (RL) is an iterative process between an agent and the
environment. The agent receives information about the environment, along with a
reward for the previous action, and then decides on a new action that will
affect the environment. Over time, the agent learns how to perform actions that
will result in the largest cumulative reward. 


\subsubsection{Actions, States, Transitions, Rewards, and Policies}
The action space, $\mathcal{A}$, is the set of all possible actions, $a$, that
can be made by an agent ($a \in \mathcal{A}$). The observation space,
$\mathcal{S}$, consists of information concerning the state $s$ of the
environment that can be used by an agent in deciding what action to take ($s \in
\mathcal{S}$). A transition function, $T(s,a,s')$, is the probability that
action $a$ from state $s$ leads to state $s'$. The output of a reward function,
$R(s,a,s')$, can be positive or negative and is represented by a real number. It
allows for deciding an action based on a decision, which may or may not move an
agent closer to a goal. A policy, $\pi$, is a function that makes a
probabilistic decision on what action to take given an observation and
previously known information about the environment. Formally,
\begin{equation*}
  \pi \colon \mathcal{A}\times \mathcal{S}\rightarrow \mathbb{R}, \quad\pi(a,s)=p(a\,|\,s).
\label{eq:policy_definition}
\end{equation*}

In general, we seek to maximize the expected discounted sum of rewards for a
policy $\pi$, i.e.,
\begin{equation*}
  Q_{\pi} (s_t,a_t) = \mathbb{E}[R_{t+1} + \gamma R_{t+2} + \gamma^2 R_{t+3} + \ldots |s_t,a_t],
\label{eq:expected_reward}
\end{equation*}
where $0 \le \gamma \le 1$ is called the discount factor. The discount factor
determines the present value of future rewards, i.e., a reward received $t$ time
steps in the future is worth only $\gamma^{t-1}$ times what it would be worth if
it were obtained immediately. If $\gamma = 0$, then the agent is only concerned
with maximizing immediate rewards. As $\gamma$ approaches 1, future rewards are
taken into account more strongly. 

\subsubsection{Epsilon-Greedy Policy}
To decide what action an agent should take, a common technique is to employ an
epsilon-greedy policy. Concretely, 
\begin{equation*}
  \pi_{\epsilon}(s,a) = 
  \begin{cases}
    1 - \epsilon + \frac{\epsilon}{|\mathcal{A}(s)|} \hspace{4mm} \text{if } a' = \argmax_{a}Q_{\pi_\epsilon}(s,a)\\ 
    \frac{\epsilon}{|\mathcal{A}(s)|} \hspace{16mm} \text{otherwise}, 
  \end{cases}
\label{eq:epsilon-greedy}
\end{equation*}
where $\epsilon$ represents the probability to `explore' the environment. In
other words, a random action is taken to evaluate if it may provide a larger
reward than previously believed. If the agent is not exploring then it is
`exploiting' where previous knowledge is used to take an action with the highest
perceived reward. A policy in which $\epsilon = 0$ is referred to as a greedy
policy. In most RL situations, $\epsilon$ starts at 1 and decays to 0 over time.

Many RL situations are sequential, i.e., more than a single action-reward pair
is iteratively performed. In contrast, scenarios in which only a single action
is taken and then the final reward is obtained are called multi-armed bandits.
Since this research explores the NBV using just a single view, it can be modeled
as a multi-armed bandit problem.

\section{Problem Statement}
\label{sec:problem_statement}
Let $X = \{x_0,\dots,x_{m-1}\} \in \mathbb{R}^3$ be a topological space where
$x_0,\dots,x_{m-1}$ are the points in a point cloud captured by a 3D sensor. Our
objective is to develop an online reinforcement learning methodology utilizing
persistent topological features for acquiring the NBV. To accomplish this, we
first approximate the topology of the space through a VR complex. Next, we
compute the view's information based on the Betti numbers of the point cloud
over a range of spatial resolutions and a deep reinforcement learning
action-value network is designed and trained using the output of our topological
information gain metric. The network is used to evaluate the NBV on a test set
of 3D objects, and the results are analyzed and discussed.

\section{Approach}
\label{sec:approach}
To find the NBV, we interpret RL actions as sensor poses. More specifically, we
model the action space as the set of all possible poses in $\mathbb{R}^6$. Since
a 3D point cloud is an unordered set of points, feeding it directly into the
network does not provide any meaningful information on its own. Instead, we
interpret the observation space as the Betti numbers of the topological
filtration of the initial view's point cloud.
We obtain the information gain for the NBV as follows. First, a topological
filtration is performed by building a VR complex at multiple spatial resolutions
on the input point cloud. Next, the Betti numbers calculated for each resolution
are algebraically combined to provide a single value that represents topological
characteristics in the observed manifold of the object. 

Intuitively, when observing a singular object {\em minimizing} $\beta_0$ will
result in an improved view by ensuring that disjoint sections become connected
through an additional view. This also minimizes the likelihood of incorporating
prior views. These prior views can have higher levels of noise outside of the
observed object thus causing the values of $\beta_0$ to increase. Conversely,
{\em maximizing} $\beta_1$ will lead to an increase in high-detailed sections of
the object manifold such as concave sections, corners, and holes. 

Due to feature scaling and the irregular density of a point cloud, a single
$(\beta_0,\beta_1)$ pair may not always provide maximal information. For this
reason, a filtration records multiple $(\beta_0,\beta_1)$ pairs at an increasing
range of distances. Thus, we define the value of a view with topological
attributes obtained at distance $r$ as 
\begin{equation}
  V = \alpha \sum_{r\in D} \beta_1(r) - (1 - \alpha)\sum_{r \in D} \beta_0(r),
\label{eq:value}
\end{equation}
where D is the set of all neighborhood point radii. The coefficient $\alpha \in
[0,1]$ can be adjusted as necessary based on the importance of connected
components ($\beta_0$) versus holes ($\beta_1$). Given a point cloud $P_t$
acquired at time step $t$, we define the reward for the NBV at $t$ as
\begin{equation}
  R_t = V(P_{t-1} \cup P_t) - V(P_{t-1}).
\label{eq:reward}
\end{equation}
This differential reward metric attempts to minimize the number of connected
components and the number of noisy areas with missing data while maximizing the
number of legitimate physical holes observed in the manifold.

\section{Experiments}  
\label{sec:experiments}
In the following subsections, we present our experimental setup and results for
computing the NBV as described in Sec.~\ref{sec:approach}. All experiments were 
conducted using streaming 3D sensor data and a robotic manipulator. Our 
software, data, and models have been made publicly available to the research
community \cite{nbvvtf2020}.

\subsection{Experimental Design}  
\label{subsec:experimental_design}
\begin{figure}
\centering
\includegraphics[width=0.35\textwidth]{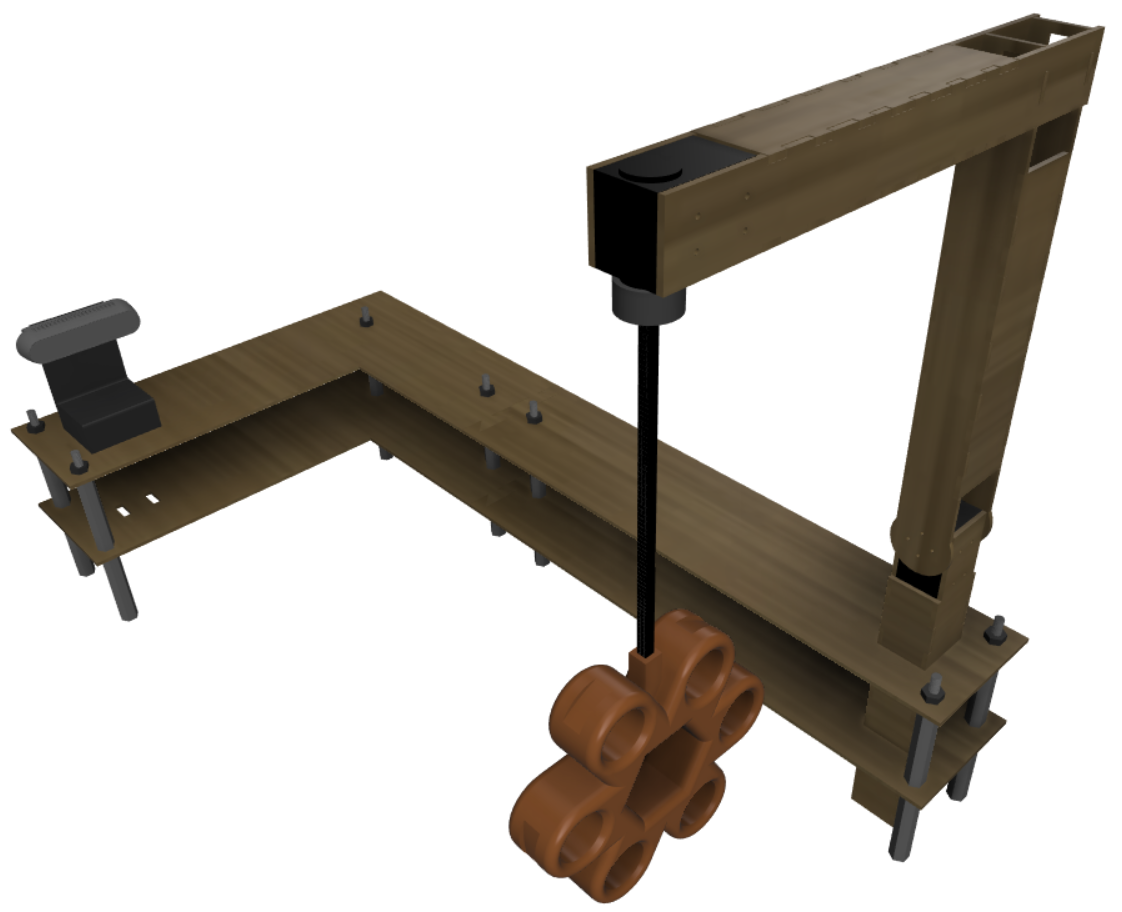}
\caption{A custom robotic manipulator designed for the NBV data collection.}
\label{fig:manipulator} 
\vspace{-4mm}
\end{figure}

To determine the NBV, a sensor must observe an object from an initial view and
then decide how to obtain an additional view based on the computed value of the
information gain. The possible positions of a sensor around an object are
6-dimensional, i.e., translation in $x,y,z$ and rotation in $\theta, \phi,
\psi$. This dimensionality may be reduced by constraining the position of the
sensor relative to the observed object such that it always points directly
towards the object, is located parallel to the ground plane, and is translated a
specific distance away from the center of the object. By making these
adjustments, the space of sensor poses can be reduced to two variables: pitch
and yaw. In a simulated environment, we would be able to position the sensor
around an object as necessary. However, sensor positioning constraints in the
real world constitute many complications. To overcome these challenges, a 2-DOF
manipulator was created to facilitate the data capture,
Fig.~\ref{fig:manipulator}. 3D point cloud data was collected with an Intel
RealSense D415 stereo camera. 
the object of observation as shown in Fig.~\ref{fig:manipulator}.

The central sensor line was made coincident with the origin of the object of
observation. Two motors (orthogonal to the sensor) were placed with axes
coincident to the central origin of the object of observation. This allowed the
bottom motor to control the pitch and the top motor to control the yaw of the
object.

The action space is all possible positions of the motors. For the manipulator, a
total of 4,096 positions are possible for each motor. Since the pitch of the
motor is not able to move a full $360^\circ$ due to the mechanics of the
manipulator, we restricted the pitch value to 2,048 positions. Although this
helped reduce the action space, it was still too large for any efficient RL
method. Therefore, we discretized the action space into buckets to further
reduce the number of possible actions to a reasonable value. For the
experiments, 20 yaw buckets and 10 pitch buckets were decided which resulted in
a final action count of 200. 

\begin{figure}
\subfloat[Object 1]{
  \includegraphics[width=0.15\textwidth]{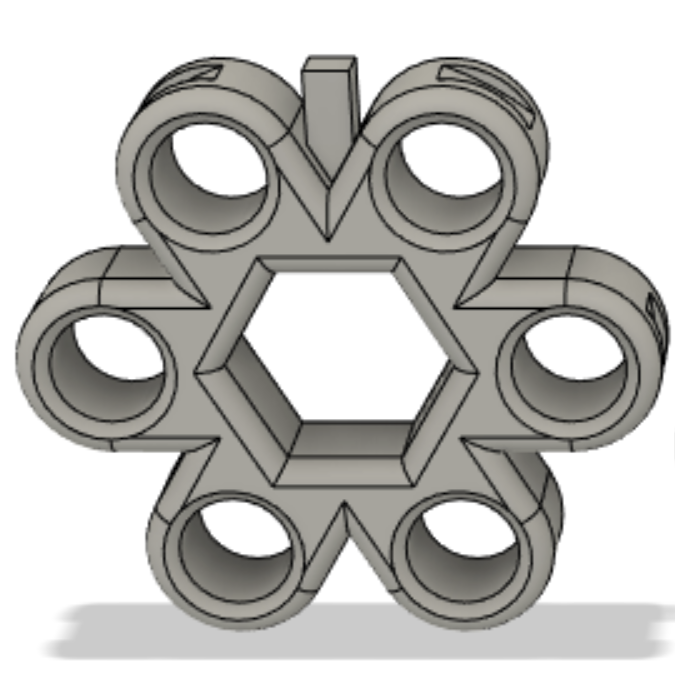}
}
\subfloat[Object 2]{
  \includegraphics[width=0.15\textwidth]{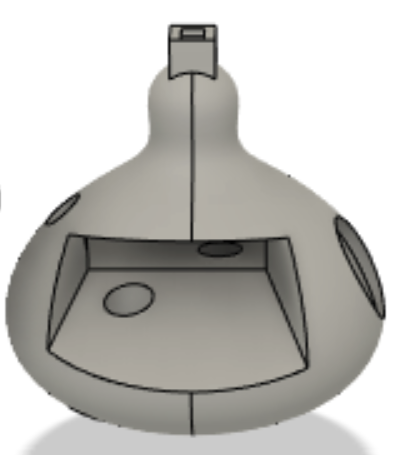}
}
\subfloat[Object 3]{
  \includegraphics[width=0.15\textwidth]{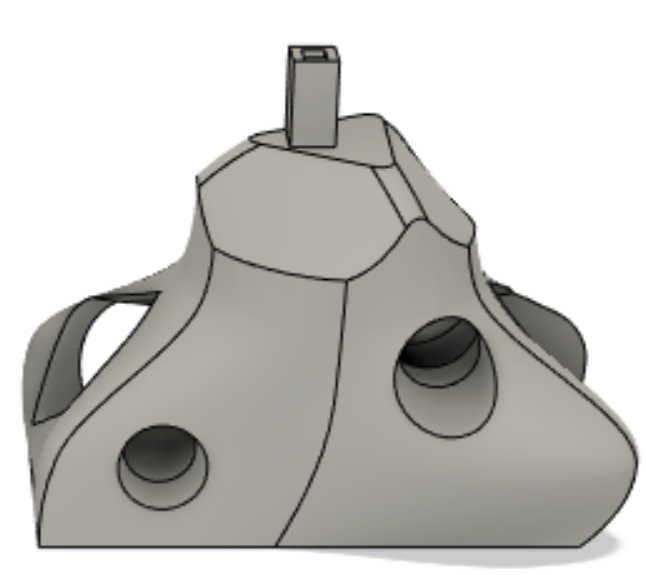}
}
\caption{Three custom feature-filled models of the seven were designed and 3D 
printed for the NBV dataset.}
\label{fig:objects} 
\vspace{-4mm}
\end{figure}

A minimum of 5,000 views were randomly sampled uniformly from the full action
space of the motors for each object. These views were saved along with a record
of the motor positions for each view. The point cloud RGB values were not used
in the experiments. We performed the data collection process for seven different
objects. Three of these items were CAD generated and included many concave
sections and holes, Fig.~\ref{fig:objects}. Three other objects were constructed
from various household objects and a toy View-Master was used for testing the
system, but not for training. Public datasets such as ShapeNet do not include 
data holes, noise, or missing sections from camera angles, therefore our score 
metric for the NBV would be inaccurate without adjustments to the data.

Transform/Union/Registration (TUR) is a software application written for our
experiments to combine point clouds. Given multiple point clouds with yaw/pitch
values, TUR will iteratively process each dataset. The first step for each point
cloud is to transform the points into the coordinate system of the world frame.
Next, the iterative closest point algorithm \cite{besl1992method} is used for
registration by matching the points in the world frame with all earlier obtained
point clouds under the presence of noise or manipulator bending. The last step
is to union the current point cloud to all previously processed point clouds.
TUR outputs a single point cloud file with all of the previous view information.
Due to the noise from the sensor and environment, as well as bending in the
manipulator, the unions are not always perfect. However, we observed that
similar reward values were obtained despite these issues. 

For a trainable RL policy, the simplest implementation is a value-averaging
tabular approach. In this method, a table entry is required for every action and
observation pair. However, while the action space is effectively discretized,
the observation space is surprisingly large. We observed values of $\beta_0$ and
$\beta_1$ greater than 200 in many samples. Based on our initial observations,
we designed a dense multilayer neural network for deep reinforcement learning.
The network takes the role of the tabular system with observations as input and
values as output, i.e., one value for each action. These values are then used to
decide what action to take. The advantage is that the observation space, given
as input to the neural network, does not need to be discretized.

We built a VR complex over the registered point cloud data via a multistep
filtration to obtain a set of $(\beta_0,\beta_1)$ pairs using the approach
described in \cite{beksi2019topology}. Owing to the computational complexity of
calculating the Betti numbers, a single RL step takes approximately 20 seconds
on a high-end workstation. We reduced the runtime by hashing and tabulating the
Betti numbers of observed viewpoints for future use. Roughly 15 minutes was
required to create the hash table of Betti values for the entire action space of
individual initial views for a single object.

Since the coordinates of the object are in an absolute world frame, the action
taken by the NBV is highly dependent on the previously observed location.
Therefore, the initial yaw and pitch values need to be known to determine the
NBV. For this reason, we include the starting yaw and pitch in the observation
space in addition to the pairs of Betti numbers. 

We note that for the observations of unioned views, the cardinality of the space
of unioned-views can be further reduced. For example, when $V_{t+1} = V_t$ the
information gain is zero and the Betti numbers remain the same (i.e., the point
cloud remains unchanged). Moreover, the union operation is commutative. Thus,
our hashing approach identifies these duplicate cases and greatly reduces the
space of unioned-views from $200 \times 200$ to $\binom{200}{2}$ = 19,900. 

\subsection{Experimental Results}
\label{subsec:experimental_results}
A neural network was trained with 2 dense layers of 64 nodes and 128 nodes
respectively, with a leaky ReLU activation function \cite{nair2010rectified}.
The output layer consisted of 200 nodes with a linear activation function.
Dropout, batch normalization, and $L_2$ regularization were utilized to prevent
overfitting and object-remembrance. To calculate the Betti numbers and the value
of a view \eqref{eq:value}, our model used a three-step filtration with the
following radii: 0.002, 0.003, 0.004. These radii were chosen from observations
of the point cloud density given the camera model and distance from the object.
Based on our observations of consistently higher values for $\beta_1$ compared
to $\beta_0$, an $\alpha$ of 0.15 was chosen to give the features approximately
equal weight. 

The training was completed in batches of 64 samples where the initial view and
object were randomly sampled. The initial observations were referenced from the
generated hash table. All future possible actions were then obtained by the
union of the observations. The network was trained with the reward labels 
\eqref{eq:reward} across all output action nodes. After 512 batches of random
initial views and objects, a test run of 512 batches was completed using
$\epsilon = 0$. The mean reward for the test batch was recorded and this
train/test cycle was completed 1,000 times. 

\begin{figure}
\begin{tabular}{ccc}
  \includegraphics[width=0.25\linewidth]{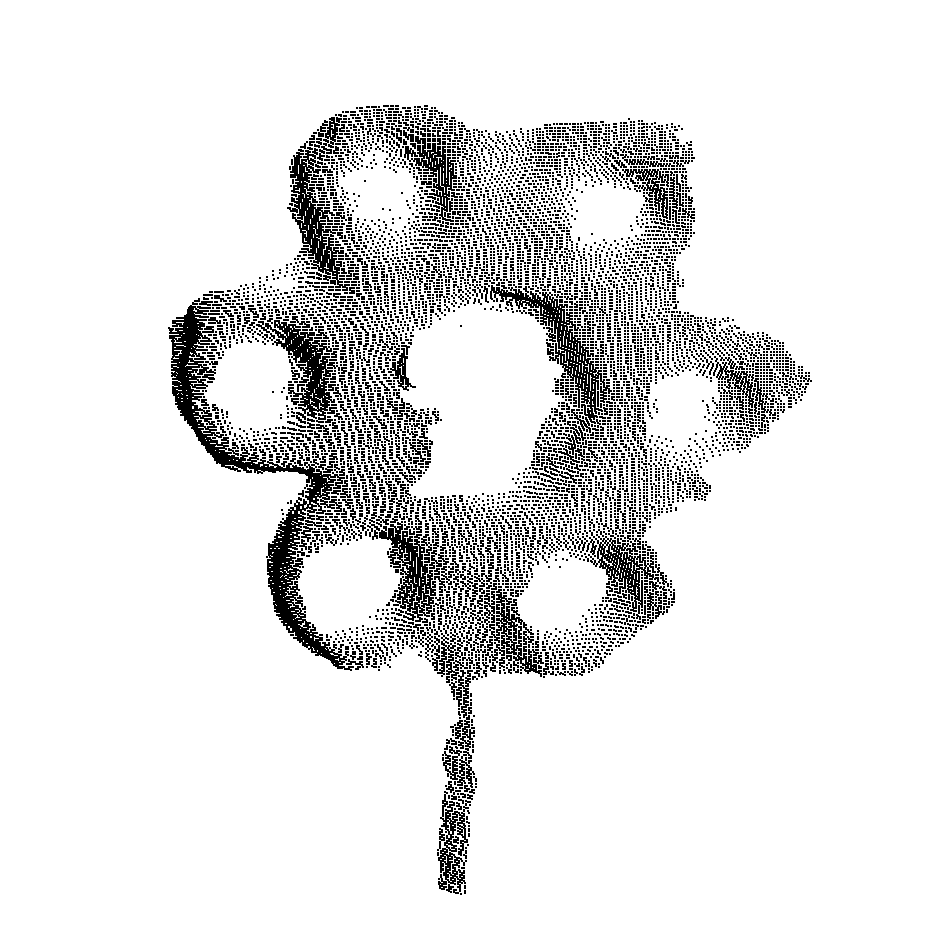} &
  \includegraphics[width=0.25\linewidth]{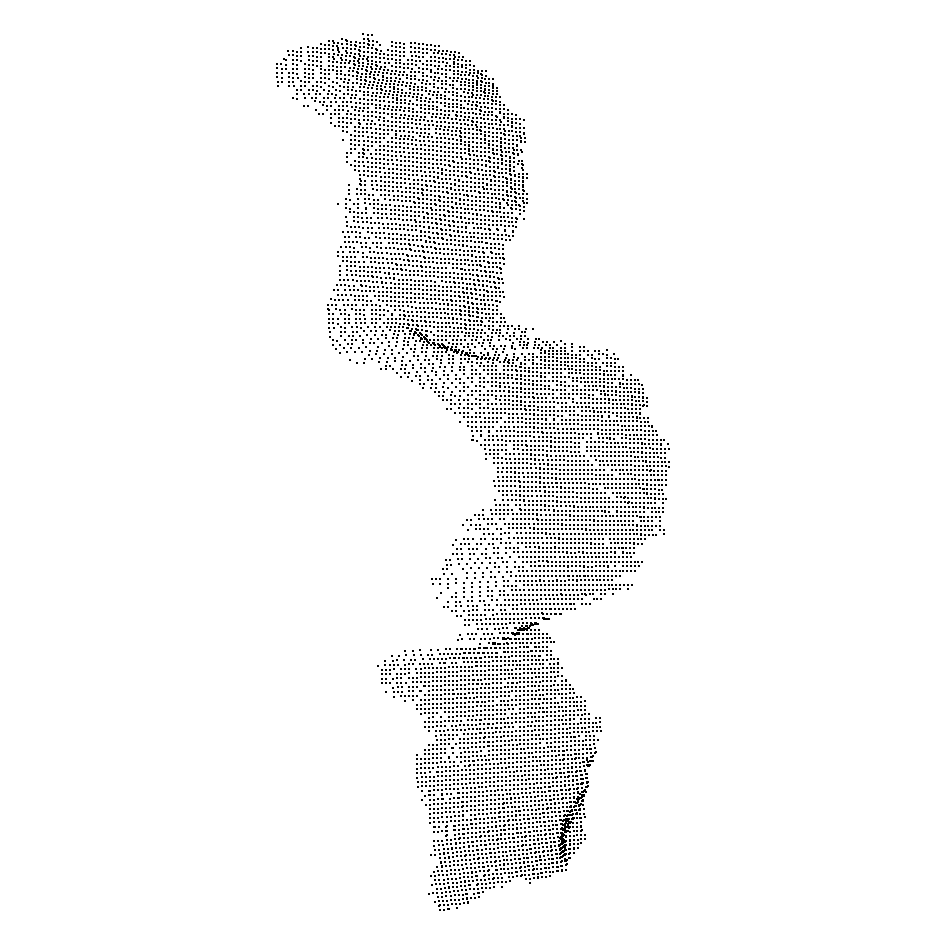} &
  \includegraphics[width=0.25\linewidth]{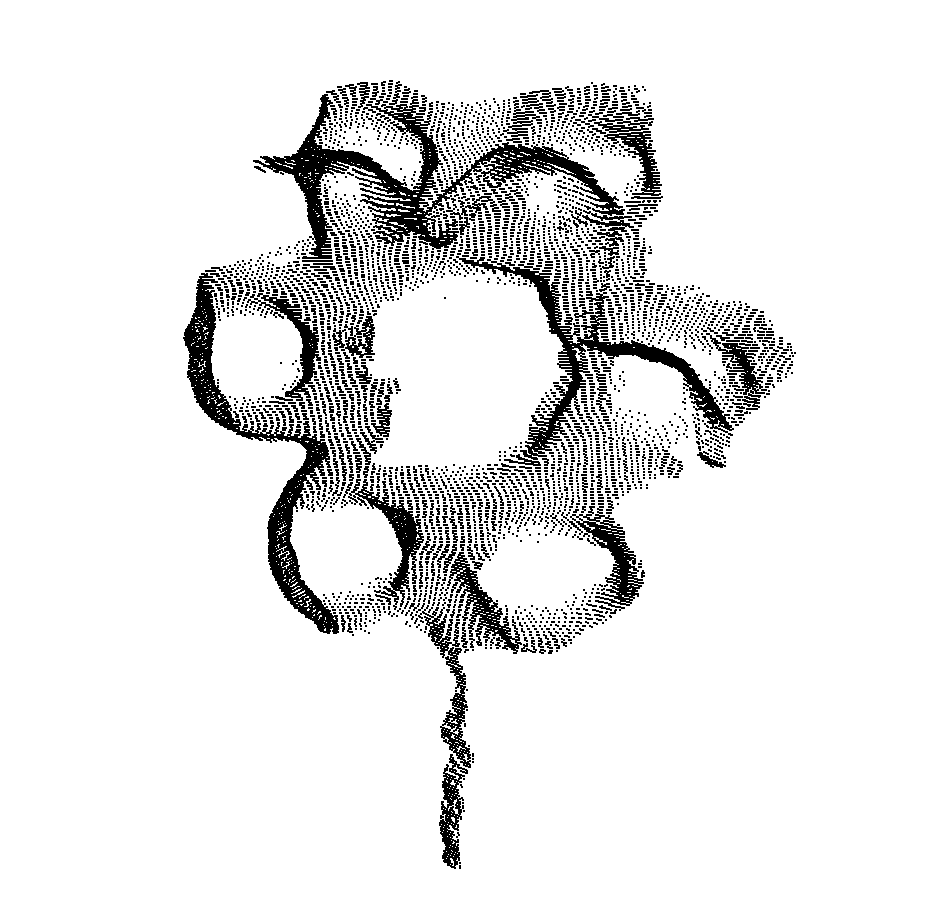}\\
  $V(P_{t-1}) = -4.5$ &
  $V(P_t) = -2.5$ &
  $V(P_{t-1} \cup P_t) = 9.7$\\
  \includegraphics[width=0.25\linewidth]{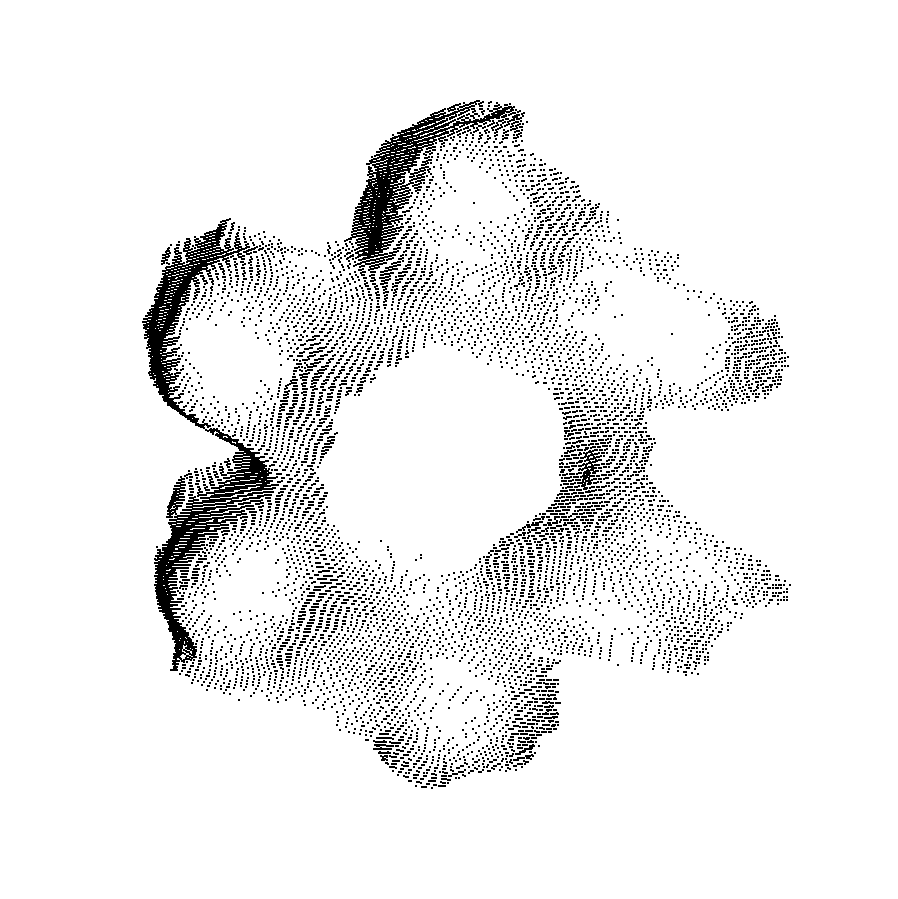}  &
  \includegraphics[width=0.25\linewidth]{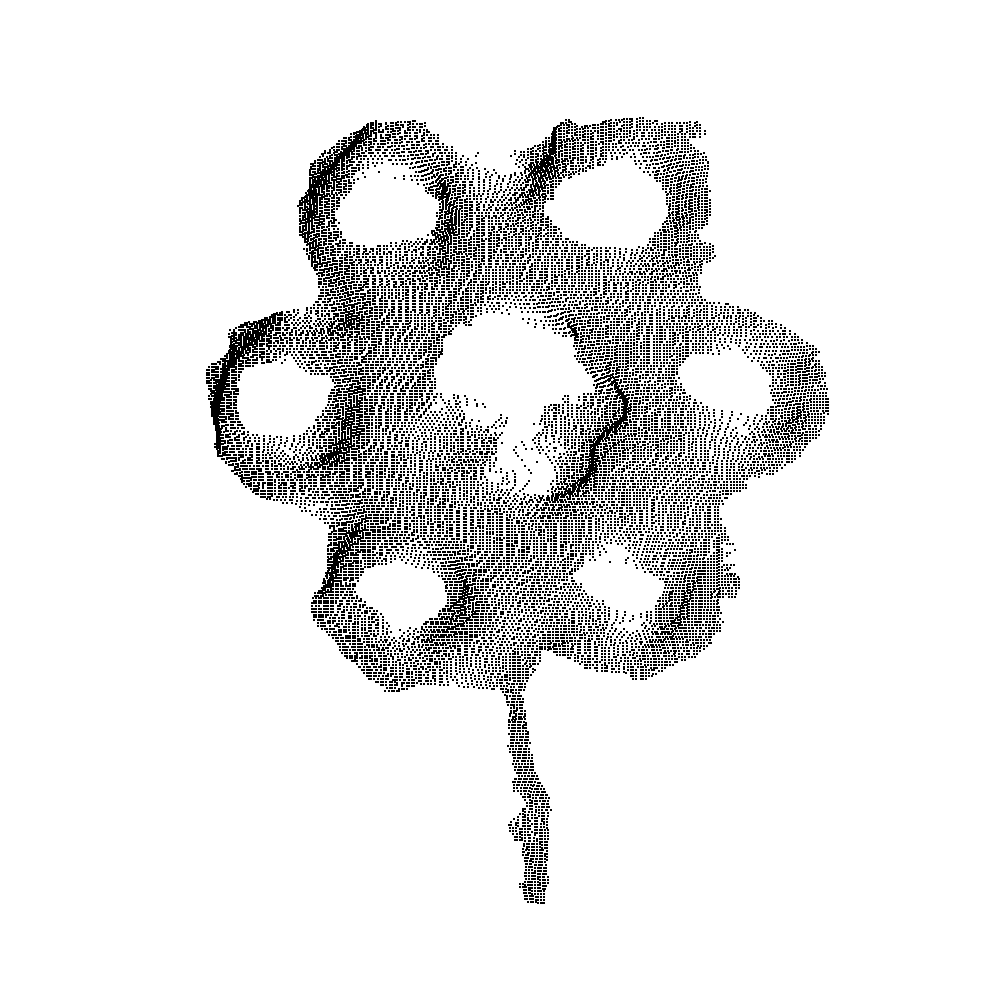} &
  \includegraphics[width=0.25\linewidth]{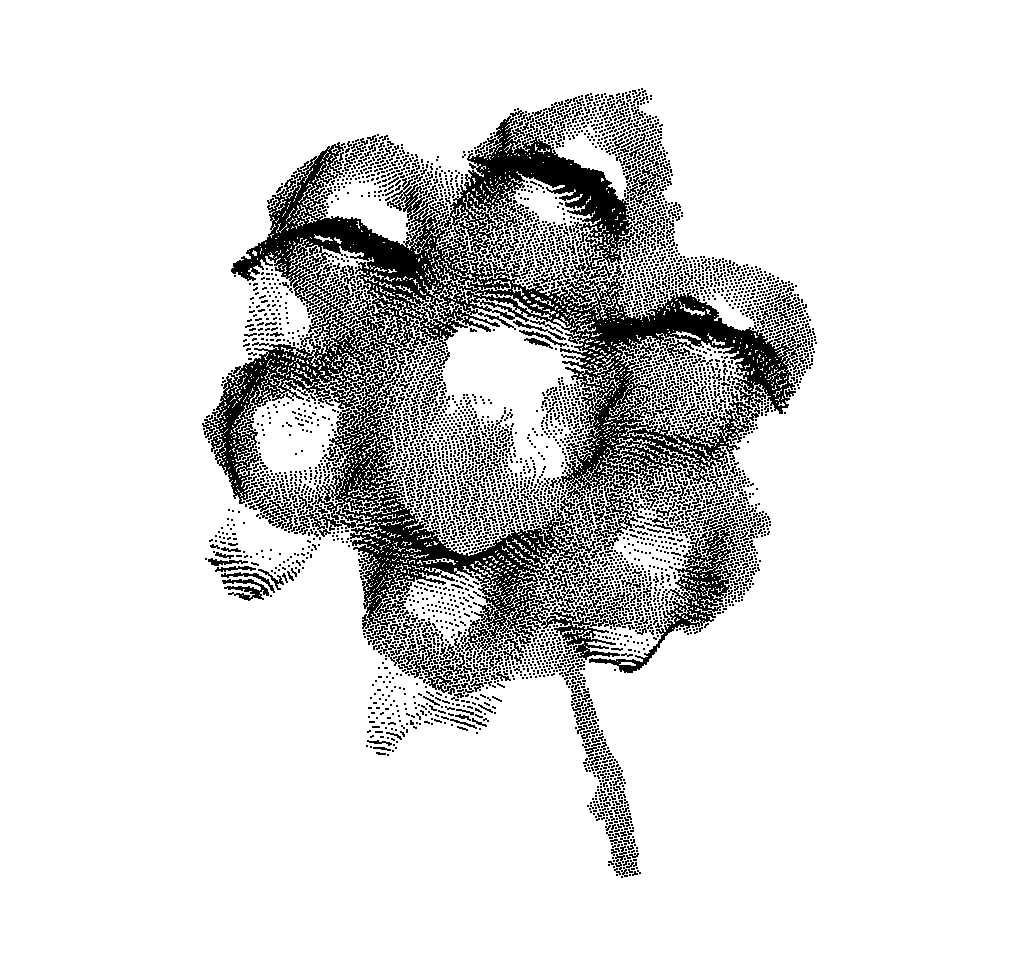} \\
  $V(P_{t-1}) = -5.3$ &
  $V(P_t) = -4.7$ &
  $V(P_{t-1}\cup P_t) = 14.2$
\end{tabular}
\caption{An example of taking the union of two low-reward values resulting in 
a higher reward value.}
\label{fig:unioned_views}
\vspace{-4mm}
\end{figure}

With the reward value defined in \eqref{eq:reward}, various situations may be
observed from the union of succeeding views. As shown in
Fig.~\ref{fig:unioned_views}, two views with individually low reward values can
be unioned together to produce a relatively high value. This indicates that the
expected reward of a new view is not as simple as adding reward values of the
individual views but instead may sometimes provide unexpected results for what
may be considered a good view.

\begin{figure}
\centering
\includegraphics[width=0.48\textwidth]{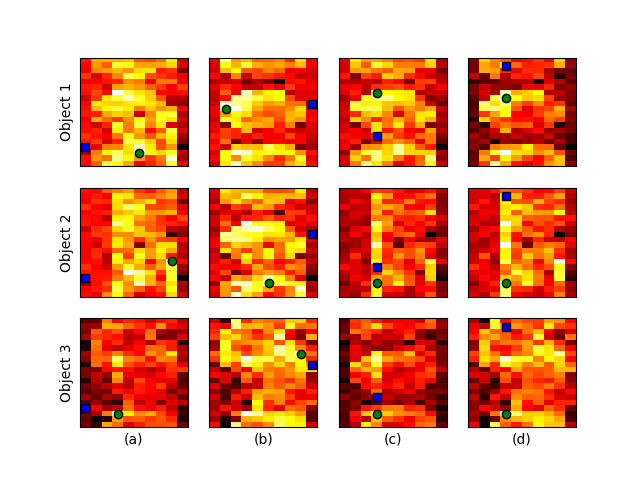}
\caption{Examples of action-value heatmaps with 
$(Y=yaw \in [0,19], X=pitch \in [0,9])$ for the three objects shown in 
Fig.~\ref{fig:objects}, and four (a)-(d) random beginning views. The blue 
squares represent the initial views while the green circles are the NBV. Lighter 
colors in the heatmap correspond to higher reward values.}
\label{fig:heatmap} 
\end{figure}

By visualizing how the rewards are distributed we can gain a better
understanding of the training process. To do this, we graph the action space of
the view in two dimensions and plot the predicted NBV reward values of the
trained model as a heatmap. We then visualize the heatmap of the final trained
model on each object with different starting views, Fig.~\ref{fig:heatmap}
(a)-(d). The initial views are indicated by blue squares in each plot and the
NBVs (i.e., $\argmax_a R(s,a)$) are illustrated by green circles. Lighter colors
in the heatmap indicate higher reward values, representing views that are
predicted to provide a large information gain. 

From Fig.~\ref{fig:heatmap} we see that reward values cluster into regions of
both optimal and negative views. For example, for the pair $(yaw=17, pitch=3)$,
42\% of the test cases consider this point to be the NBV. Pitch values in the
interval $[0,2]$ consistently show negative rewards, as does a pitch value of 9.
On the other hand, a pitch value of 3 shows a light-colored column in most of
the object heatmaps thus considering it to be a good view. Repetitive regions
can be seen in the heatmaps for object 1(a) and 1(b) in Fig.~\ref{fig:heatmap}.
These repeating areas are also present in the heatmap of object 3(b) and 3(c),
albeit slightly shifted. The shift can be explained as the NBV is contingent on
relative viewing angles. Note that the yaw coordinate is a full 360\degree,
i.e., each image wraps around vertically. Two primary reasons for these regions
are the following: since we consider angles that are relative to the object
frame each of the objects has a $90^\circ$ and $180^\circ$ symmetry, and certain
positions of the manipulator occlude the view in many cases. 

\begin{figure}
\centering
\begin{tabular}{m{1.9cm}|c|c|}
  \cline{2-3}
  & Initial View & NBV\\
  \cline{2-3}
  & &\\ 
  Object 1(d) & \includegraphics[width=0.25\columnwidth,height=0.25\columnwidth]{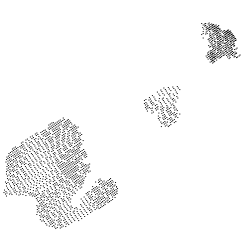} &
  \includegraphics[width=0.25\columnwidth,height=0.25\columnwidth]{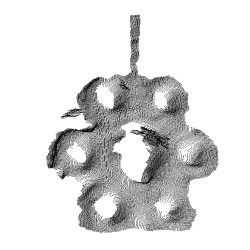}\\
  \cline{2-3}
  & &\\
  Object 2(a) & \includegraphics[width=0.25\columnwidth,height=0.25\columnwidth]{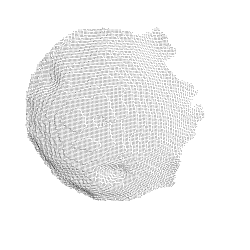} & \includegraphics[width=0.25\columnwidth,height=0.25\columnwidth]{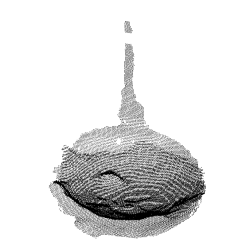}\\
  \cline{2-3}
  & &\\
  Object 3(b) & \includegraphics[width=0.25\columnwidth,height=0.25\columnwidth]{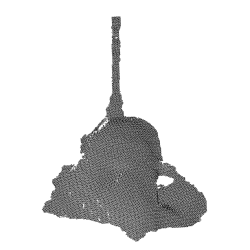} & \includegraphics[width=0.25\columnwidth,height=0.25\columnwidth]{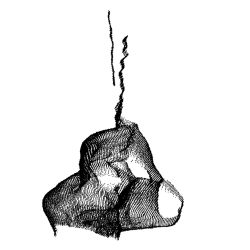}\\
  \cline{2-3}
\end{tabular}
\caption{Examples of initial views and the NBVs for the objects in 
Fig.~\ref{fig:heatmap}.}
\label{fig:nbv_examples} 
\vspace{-2mm}
\end{figure}

\begin{figure}
\centering
\begin{tabular}{|c|c|}
  \cline{1-2}
  Initial View & NBV\\
  \cline{1-2}
  &\\ 
  \includegraphics[width=0.25\columnwidth,height=0.25\columnwidth]{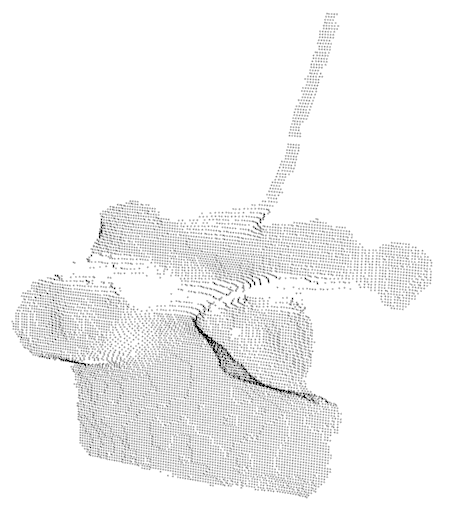} &
  \includegraphics[width=0.25\columnwidth,height=0.25\columnwidth]{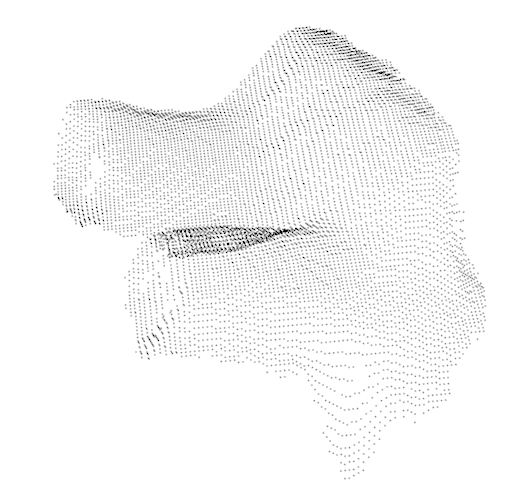}\\
  \cline{1-2}
\end{tabular}
\caption{An example of an initial view and the NBV for the test object.}
\label{fig:test_example} 
\vspace{-4mm}
\end{figure}

\begin{figure*}
\centering
\includegraphics[width=0.90\textwidth]{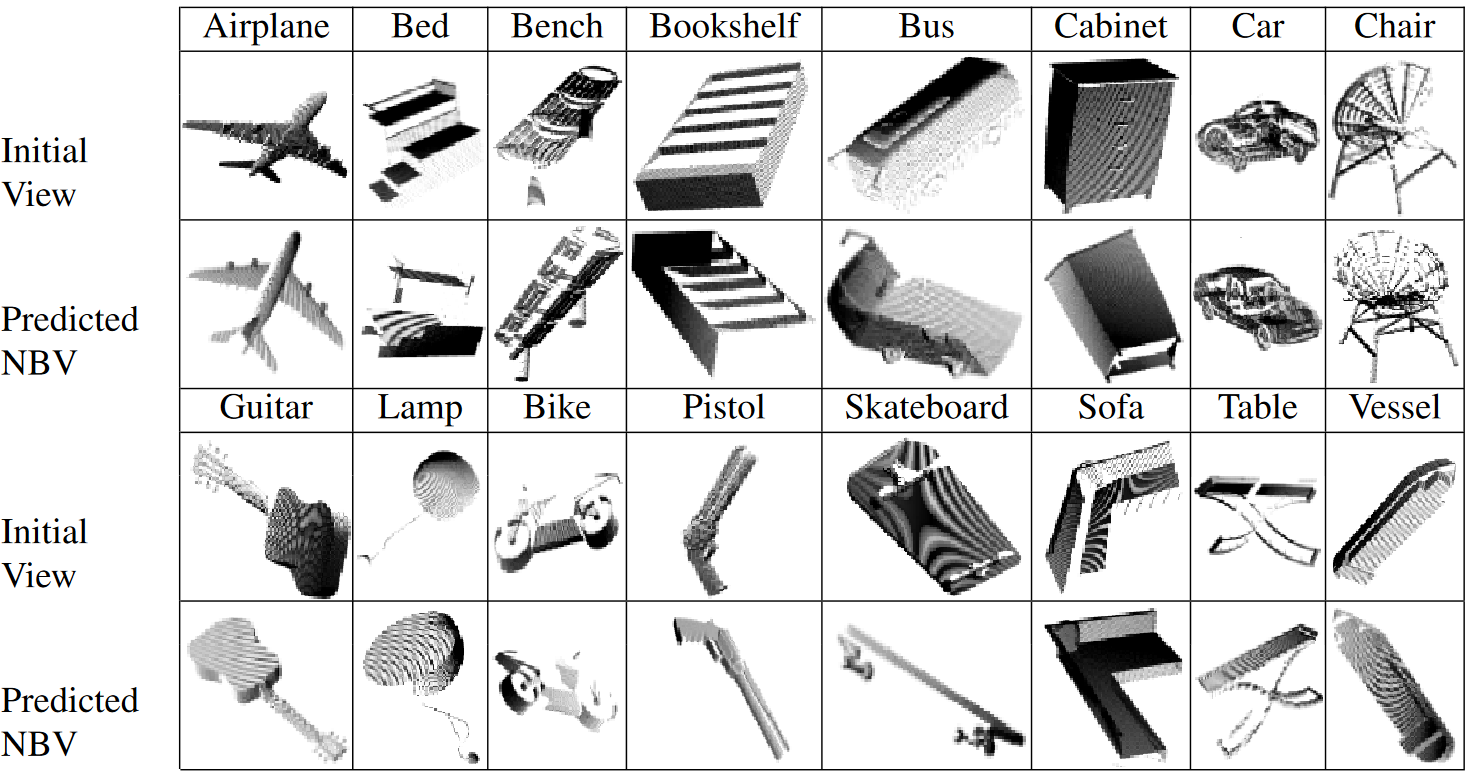}
\vspace{-2mm}
\caption{Examples of initial views and the NBVs for various ShapeNet
\cite{chang2015shapenet} objects.}
\label{fig:shapenet} 
\vspace{-4mm}
\end{figure*}

The point clouds of objects 1(d), 2(a), and 3(b) are shown in
Fig.~\ref{fig:nbv_examples}. For object 1(d), the initial view provides little
detail and few points due to the obstruction of the manipulator. The predicted
NBV is a partially-sideways front view that allows the two disjoint sections to
join together into a single cluster when merged with the original pieces. For
object 2(a), the spherical bottom of the object is initially observed. The
prediction for the NBV is a side view of the object, and a larger representation
of the object is realized when the union is performed. The inconsistency of the
transform can be attributed to the bending of the manipulator. For object 3(b),
the initial view is a full side view. The NBV prediction is of a partial side
view of the object. Given these two views, there is enough overlapping
information regarding the sides of the object in each side view that they can be
unioned as one view. For all these examples, the NBV provided additional
viewpoints that allowed intersecting information between the initial and ensuing
views. This resulted in cohesive viewpoints of increasing detail. It shows that
our topological information gain metric and RL approach correctly identify
high-reward views, and it effectively increases the point cloud density around
probable features.

Since the previous tests were completed using training data, a seventh
real-world object was used for unbiased testing afterwards. The information gain
values were calculated from the initial and union Betti numbers as with the
previous objects, but these values were not used when training the RL model. The
NBV returned from our RL model on an example initial view can be seen in
Fig.~\ref{fig:test_example}. In this example, the initial view is rather good.
Yet, there is an entire missing section on the back side of the object in
addition to holes underneath and above the eye section due to being coplanar
with the camera. In the predicted NBV, a lower angle is obtained which allows
many of the missing sections to be cleanly filled and provides finer detail
around the edges and curves.

To further test the model after training, 16 ShapeNet objects from the same
categories as in \cite{zeng2020pc} were placed into a simulated Gazebo
\cite{gazebo} environment. A Robot Operating System \cite{ros} model for the
simulated manipulator was created and a virtual Intel RealSense camera with
parameters matching the physical model was used. With this system, the data
collection process was repeated with the ShapeNet objects and the NBV was
predicted using the trained model. As shown in Fig.~\ref{fig:shapenet}, the
trained NBV model tends to favor viewpoints that will partially overlap the
initial viewpoint. This creates a combined manifold that attempts to better
cover feature-filled areas of the objects such as holes and edges. In
Fig.~\ref{fig:shapenet}, the initial views were randomly chosen in a uniform
manner. In addition, the predicted NBV images consist of only the second view's
point cloud, as opposed to the unioned view. This allows for an easier
visualization of the views.

To better quantify the decisions of the NBV on the ShapeNet objects, ten initial
views from each object were evaluated with our trained model. The resulting
angular difference between the initial view and the NBV were calculated. Of the
160 tests, the mean angular difference was $61.8^\circ$ with a standard
deviation of $19.2^\circ$. These results fit with the intuitive beliefs of our
metric. Very small angles will not correctly fill in holes of missing points due
to occlusions. Large angles (e.g., $180^\circ$) tend to increase the number of
connected components and sections with missing points resulting in a lower value
from our metric. 
Thus, viewpoints are chosen that are neither too far nor too close to the
initial view which allows for probable feature-filled sections to be better
realized. 



\section{Conclusion and Future Work}
\label{sec:conclusion}
This paper introduced an online RL framework, modeled as a multi-armed bandit
problem, for streaming 3D point cloud data. Moreover, we proposed a unique
topology-based information gain metric to determine the NBV for a noisy 3D
sensor. The metric focuses on topological features, such as holes and concave
sections, and computes information gain by combining disjoint sensor viewpoints.
Experimental results show that our system can help determine the placement of an
object held by a robotic manipulator. As part of this work, a labeled dataset of
3D objects, a CAD design for a custom robotic manipulator, and software for the
transformation, union, and registration of 3D point clouds given their viewpoint
coordinates have been publicly released. 

For future work, we plan to develop and test additional salient information gain
metrics. For example, the volume of the convex hull of a point cloud is a metric
that may be used for the RL observation space. The potential advantage of using
the volume of the convex hull is that the added information would complement our
existing topological metric. Another metric that could be used is the density of
a point cloud in regions that are colinear to the view of the sensor. The
benefit of such a metric is that views will be chosen that have relatively low
density hence indicating a lack of features. The NBV will union together views
from these sparse locations thus increasing the overall density of points. 

\vspace{-2mm}
\section*{Acknowledgments}
This material is based upon work supported by the National Science Foundation
through grants \#IIS-1724248 and \#IIS-1947851, and a Department of Education's
Graduate Assistance in Areas of National Need grant \#1261801280. We thank
Joseph M. Cloud for helpful discussions and comments on this work. 

\bibliographystyle{IEEEtran}
\bibliography{IEEEabrv,learning_the_next_best_view_for_3d_point_clouds_via_topological_features}   
\end{document}